# SAD-GAN: Synthetic Autonomous Driving using Generative Adversarial Networks

**Arna Ghosh**[*]
Department of Electrical Engineering
Indian Institute of Technology
Kharagpur, WB 721302. India.
`arnaghosh@iitkgp.ac.in`

**Biswarup Bhattacharya**[*]
Department of Electrical Engineering
Indian Institute of Technology
Kharagpur, WB 721302. India.
`biswarup@iitkgp.ac.in`

**Somnath Basu Roy Chowdhury**[*]
Department of Electrical Engineering
Indian Institute of Technology
Kharagpur, WB 721302. India.
`brcsomnath@ee.iitkgp.ernet.in`

## Abstract

Autonomous driving is one of the most recent topics of interest which is aimed at replicating human driving behavior keeping in mind the safety issues. We approach the problem of learning synthetic driving using generative neural networks. The main idea is to make a controller trainer network using images plus key press data to mimic human learning. We used the architecture of a stable GAN to make predictions between driving scenes using key presses. We train our model on one video game (Road Rash) and tested the accuracy and compared it by running the model on other maps in Road Rash to determine the extent of learning.

## 1 Introduction

Self-driving cars are one of the most promising prospects for near term artificial intelligence research. Autonomous driving is a well-established problem and the use of large amounts of labeled and contextually rich data to solve the problems of road detection and prediction of vehicle parameters like accelerator, clutch and brake positions have already been explored [5]. However, a major challenge is a dataset that is sufficiently rich to cover all situations as well as different conditions. A solution proposed to aid the issue is the use of synthetic data along with natural data to train the system [12].

Driving is a task that demands complicated perception and controls tasks which are intricately linked to each other. The technology to correctly solve driving can potentially be extended to other interesting tasks such as action recognition from videos and path planning in robotics. Vision based controls and reinforcement learning had recent success in the literature [6], [9], [14], [8] mostly due to (deep, recurrent) neural networks and unbounded access to world or game interaction. Such interactions provide the possibility to revisit states with new policies and to simulate future events for training deep neural network based controllers.

To understand the controls or the basic plant model of the vehicle, two methods are possible - prepare a simulator and study the response of the plant to different inputs or to learn to simulate. The use of Generative Adversarial Networks [4] for the same has been explored in [13] and acts as a good motivation for the work presented.

---

[*]Equal Contribution

This paper explores the idea to use generative networks for predicting the next state of the vehicle, or more accurately the next camera feed from a camera mounted on a vehicle given a motion vector. The use of synthetic idea for the same is novel as per the author's knowledge. Once a satisfactory generative network is obtained, the generator can be used to generate a variety of images to explore the next steps similar to that used in Atari games [2], thus building an alpha-beta pruned game tree. Each action is scored according to how far down the game tree is the vehicle predicted to be "safe".

The authors believe that the use of synthetic data from one game can be used to train the entire network and then the network can be used on a different game to predict performance. This is because the generator trained signifies the plant-plus-sensor loop. But the controller is trained using reinforcement learning with the reward and punishment defined as a function of the layer when the move becomes unsafe. Therefore, changing the plant may result in a temporary fall in accuracy but will eventually lead to better results as the system plays the game.

The present methodology presented here deals with prediction of images that are marked as "safe" or "unsafe". But the network could be modified to achieve the task in a latent space. Since the present method allows easier visualization, the authors believe it would be easier to establish the concept and then look into optimization techniques to make it usable in present day techniques.

## 2 Methodology

Our entire work in synthetic driving has been performed in the following sequential order namely labeled data collection, generative network, predictive convolutional network, training and testing process.

### 2.1 Data Collection

For the purpose of the paper, we used frames from the popular racing game *Road Rash* [1]. Each image consisted of the driver's bike, the surrounding environment details including the road, sky, traffic, fences, grass etc. Also, we required the key press information as can be observed in Figure 1. For the purpose of tracking key press information, we created a keylogger software to log the keys pressed and take a snapshot of the image on the screen at that time. This was done over multiple races to get as many pictures as required. The average time to collect data per race is 2 minutes and the average number of usable images collected per race can be as high as $500$.

### 2.2 Architecture

Figure 1: Generator GAN model

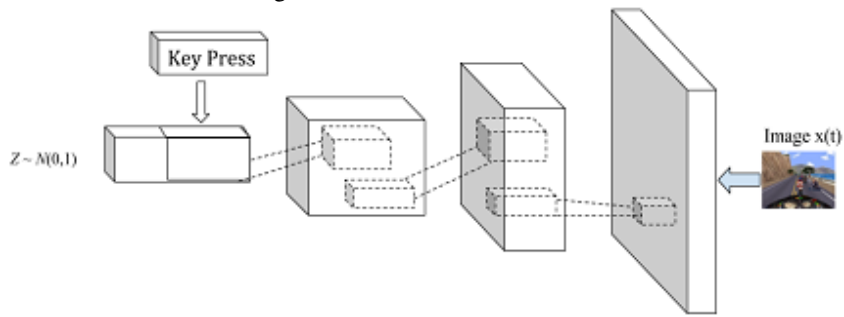

In order to implement the concept of synthetic driving we have used the standard architecture of DCGAN. The input to the DCGAN is the labeled image of a driving scene and the corresponding key press label. The architecture of a stable deep convolutional generative adversarial network (DCGAN) is utilized [10]. The first layers of the stable DCGAN consist of a convolutional neural network performs strided convolution unlike CNNs which performs spatial sampling. The fully connected layers are absent in DCGAN. The features extracted from the highest convolutional layer form the input of the generator and output of the discriminator. In order to stabilize the learning process, batch normalization is applied to the discriminator input. Batch normalization is used for normalizing the input to have a zero mean and unit variance. The architecture for the generator of the DCGAN is shown in Figure 1.

The generator and discriminator architecture are inspired from [11]. The generator receives the input image at current time x(t) instant and the key press (up, left or right) at that time in the game. The generator then tries to simulate the driving scene at the next time instant x(t+1). The discriminator receives as input the actual driving scene at x(t+1). The discriminator has convolution network followed by leaky ReLu. The discriminator passes the input image through convolution layers to create a feature map. The discriminator then compared it with the generator output to cross-verify and aid the training process of the generator. The discriminator network architecture is shown in Figure 2.

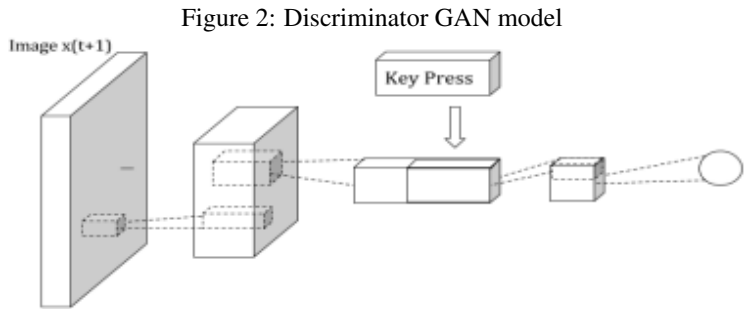

Figure 2: Discriminator GAN model

The training of the neural network is performed on a standard AlexNet architecture [7]. The training occurs by taking two images x(t) and x(t+1) as inputs to the neural network. After the convolution layers the tensors produced for each image is concatenated and labeled as the key press used to achieve the transition. The architecture consists of five convolutional layers in which the first two and final convolutional layers are followed by max-pooling. Batch normalization of each ReLu layers are performed. The convolutional layers are followed by three fully connected layers where the dropout probability is set at $0.5$. During actual driving simulation the image of a present driving scene is used to generate the scene x(t+1) using the architecture of a stable DCGAN. The input image x(t) and the generated image x(t+1) is then fed to the neural network architecture [7] to predict the key press from the trained network. For optimization purposes we have used stochastic gradient descent using momentum vector $(0.9)$ and trained the model in small batches. The full architecture of the network is shown in Figure 3.

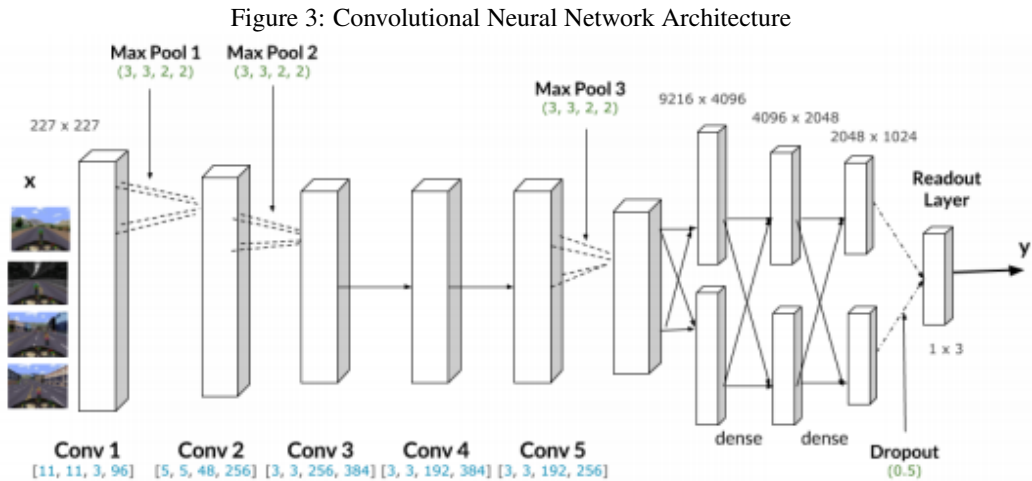

Figure 3: Convolutional Neural Network Architecture

### 2.3 Training and Workflow

The authors use the dataset collected to train a generator network to predict images given an image and a key press. The discriminator is trained to distinguish between generated images and images from the dataset. After obtaining a sufficiently efficient generator, the generator network is deployed in action to predict all three images from a given image. The three images are the images that should result from left, up and right key press from the present situation.

The three images thus generated are classified as "safe" or "unsafe". For this task, the authors train a simple network from the above dataset of images collected from Road Rash. The network is trained to predict the key press the user would have chosen from a given scene. So for a given image, the network is supposed to predict left, right or up as the key press. The results for this are significantly motivating, however it depends on the game platform. So for each game, the network needs to be trained to establish an image as "safe" or "unsafe". If the predicted key press is same as the one that generated the image from the previous level, thus indicating that going down one more level is feasible, it is marked as "safe" else "unsafe".

The metric for reinforcement learning is set as the maximum number of levels down the game tree the decision yields a safe scene. Therefore, the metric is not specified explicitly but is implicitly derived by the network itself. The safety of the vehicle is prioritized over the rank or speed at which the vehicle is moving in this strategy.

The authors believe that even humans learn to play the game in a similar fashion. Players also take into account the rank and/or speed along with this data. But in the overall sense, the players tend to implicitly predict the consequences of their action and choose the one which they believe is the safest. Also, the reinforcement strategy employed here is somewhat similar to what humans use in due course of the game. With each failure, they learn to predict the safety quotient of a situation and modify their responses accordingly to choose the situation with maximum safety quotient.

## 3 Results

An interesting move is to keep all the convolution layers constant over the generator, discriminator and the "safe" or "unsafe" labeling network. This would allow a shift to the latent space in the near future. The overall network used for prediction of "safe" or "unsafe" yielded an accuracy of $90\%$ using three classes each containing $200$ images after training the model for $25$ epochs. The training error for 20 epochs is shown below in Figure 4. The training process approximately took one hour for completion.

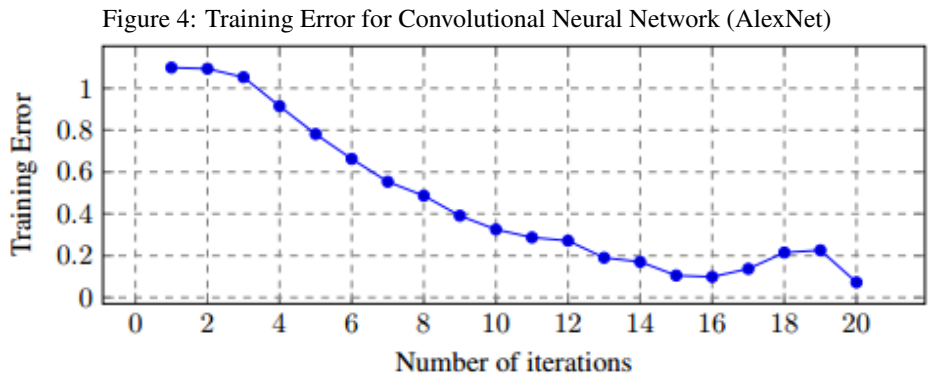

Figure 4: Training Error for Convolutional Neural Network (AlexNet)

## 4 Further Work

Presently the generator-discriminator network is being tested on data obtained from Road Rash. Further experiments would include testing on other games like Grand Theft Auto. A litmus test to this method would be to include natural images from the KITTI dataset [3] along with an ego-motion vector quantified enough to fit the architecture presented here and observe the results.

Our network doesn't rely on standard methods like object recognition or scene labeling for decision making. Once trained on well-labeled real world data over a finite period of time we can use the generator model to predict more than one image [x(t), x(t+1), ..., x(t+n)] and make decisions based on the entire cluster of images. We can also add other real world driving parameters into the system like gear, acceleration, braking etc. using an auto-encoder.

## 5  Conclusion

Our algorithm provides an insight into improving the state of the art algorithms in autonomous driving by predicting future driving scenes using generative methods. The network also trains itself according to the behavior of the driver whose data is being fed into the network. Extending this idea, the network presented can also be used in case of manual driving where it can act as a recommendation system for the driver by predicting different situations once it is trained on real world driving scenarios.

## References

[1] E. Arts. Road rash. [CD-ROM], 1991.

[2] G. Chaslot, S. Bakkes, I. Szita, and P. Spronck. Monte-carlo tree search: A new framework for game ai. In C. Darken and M. Mateas, editors, *AIIDE*. The AAAI Press, 2008.

[3] A. Geiger, P. Lenz, C. Stiller, and R. Urtasun. Vision meets robotics: The kitti dataset. *International Journal of Robotics Research (IJRR)*, 2013.

[4] I. J. Goodfellow, J. Pouget-Abadie, M. Mirza, B. Xu, D. Warde-Farley, S. Ozair, A. Courville, and Y. Bengio. Generative adversarial networks. *ArXiv e-prints*, jun 2014.

[5] Z. Halim, R. Kalsoom, and A. R. Baig. Profiling drivers based on driver dependent vehicle driving features. *Applied Intelligence*, 44(3):645–664, 2016.

[6] J. Koutník, G. Cuccu, J. Schmidhuber, and F. Gomez. Evolving large-scale neural networks for vision-based reinforcement learning. In *Proceedings of the 15th Annual Conference on Genetic and Evolutionary Computation*, GECCO '13, pages 1061–1068, New York, NY, USA, 2013. ACM.

[7] A. Krizhevsky, I. Sutskever, and G. E. Hinton. Imagenet classification with deep convolutional neural networks. In F. Pereira, C. J. C. Burges, L. Bottou, and K. Q. Weinberger, editors, *Advances in Neural Information Processing Systems 25*, pages 1097–1105. Curran Associates, Inc., 2012.

[8] S. Levine, P. Pastor, A. Krizhevsky, and D. Quillen. Learning hand-eye coordination for robotic grasping with deep learning and large-scale data collection. *CoRR*, abs/1603.02199, 2016.

[9] V. Mnih, K. Kavukcuoglu, D. Silver, A. A. Rusu, J. Veness, M. G. Bellemare, A. Graves, M. Riedmiller, A. K. Fidjeland, G. Ostrovski, S. Petersen, C. Beattie, A. Sadik, I. Antonoglou, H. King, D. Kumaran, D. Wierstra, S. Legg, and D. Hassabis. Human-level control through deep reinforcement learning. *Nature*, 518(7540):529–533, feb 2015.

[10] A. Radford, L. Metz, and S. Chintala. Unsupervised representation learning with deep convolutional generative adversarial networks. *CoRR*, abs/1511.06434, 2015.

[11] S. E. Reed, Z. Akata, X. Yan, L. Logeswaran, B. Schiele, and H. Lee. Generative adversarial text to image synthesis. *CoRR*, abs/1605.05396, 2016.

[12] S. R. Richter, V. Vineet, S. Roth, and V. Koltun. Playing for data: Ground truth from computer games. *CoRR*, abs/1608.02192, 2016.

[13] E. Santana and G. Hotz. Learning a driving simulator. *CoRR*, abs/1608.01230, 2016.

[14] D. Silver, A. Huang, C. J. Maddison, A. Guez, L. Sifre, G. van den Driessche, J. Schrittwieser, I. Antonoglou, V. Panneershelvam, M. Lanctot, S. Dieleman, D. Grewe, J. Nham, N. Kalchbrenner, I. Sutskever, T. Lillicrap, M. Leach, K. Kavukcuoglu, T. Graepel, and D. Hassabis. Mastering the game of Go with deep neural networks and tree search. *Nature*, 529(7587):484–489, jan 2016.